\pdfoutput=1 
\documentclass[11pt,a4paper]{article}
\usepackage[hyperref]{eacl2021}
\usepackage{times}
\usepackage{latexsym}

\usepackage{microtype}
\usepackage[ruled,vlined,linesnumbered]{algorithm2e}
\usepackage{amsmath,amsfonts,multirow}

\aclfinalcopy 

\title{Trajectory-Based Meta-Learning for Out-Of-Vocabulary Word Embedding Learning}

\author{Gordon Buck \\
  University of Cambridge \\
  \texttt{ghb28@cantab.ac.uk} \\\And
  Andreas Vlachos \\
  University of Cambridge \\
  \texttt{av308@cam.ac.uk} \\}

\date{}

\begin{document}
\maketitle
\begin{abstract}
Word embedding learning methods require a large number of occurrences of a word to accurately learn its embedding. However, out-of-vocabulary (OOV) words which do not appear in the training corpus emerge frequently in the smaller downstream data. Recent work formulated OOV embedding learning as a few-shot regression problem and demonstrated that meta-learning can improve results obtained. However, the algorithm used, model-agnostic meta-learning (MAML) 
is known to be unstable and perform worse when a large number of gradient steps are used for parameter updates.
In this work, we propose the use of Leap, a meta-learning algorithm which leverages the entire trajectory of the learning process instead of just the beginning and the end points, and thus ameliorates these two issues. In our experiments on a benchmark OOV embedding learning dataset and in an extrinsic evaluation, 
 Leap performs comparably or better than MAML. We go on to examine which contexts are most beneficial to learn an OOV embedding from, and propose that the choice of contexts may matter more than the meta-learning employed.

\end{abstract}

\section{Introduction}

Distributional methods for learning word embeddings require a sufficient number of occurrences of a word in the training corpus to accurately learn its embedding.
Even though the embeddings can be trained on raw text implying that an embedding for every word is obtained, in practice 
out-of-vocabulary (OOV)
words do occur in the downstream applications embeddings are used,
for example 
due to
domain-specific terminology. 
Nevertheless, OOV words are often content words such as names which convey important information for downstream tasks; for example, drug names are key
in the biomedical domain. However, the amount of downstream language data is typically much smaller than the corpus
used for training word embeddings, thus methods that rely on distributional properties of words across large amounts of data perform poorly \citep{herbelot2017highrisk}.

Researchers often assign OOV words to random embeddings or to an ``unknown'' embedding, however these solutions fail to capture the distributional properties of words. 
Zero-shot approaches \citep{pinter-etal-2017-mimicking, kim2015characteraware, bojanowski2016enriching} attempt to predict the embeddings for OOV words from their characters alone. These approaches rely on inferring the meaning of a word from its subword information, such as morphemes or WordPiece tokens used in BERT \citep{devlin-etal-2019-bert}. While this works well for many words, it performs poorly for names and words where morphology is not informative.

Given that an OOV word occurs once, the chance of a second occurrence is much higher than the first \citep{noriegas}. Hence while OOV words can be rare and not seen in training, it is reasonable to expect that a limited number of occurrences will be present in the data of a downstream application. 
Few-shot approaches \citep{garneau-etal-2018-predicting, khodak2018la,hu2019fewshot}  leveraged this to predict the embeddings for OOV words from just a few contexts, often in conjunction with their morphological information.
\citet{hu2019fewshot} proposed an attention-based architecture for OOV word embedding learning as a few-shot regression problem. The model is trained to predict the embedding of a word based on a few contexts and its character sequence. Such a model is trained by simulating OOV words in the training corpus, with their target embeddings provided by learning them
on the same corpus. As OOV words 
must have their embeddings inferred from contexts outside the training corpus, 
the authors show that using an adaptation of the model-agnostic meta-learning (MAML) algorithm \citep{finn2017modelagnostic} to adapt the model's parameters to the target domain improves the quality of the learned OOV word embeddings.

However, MAML is known to be
unstable 
due to the calculation of gradients requiring 
backpropagation through multiple instances of the model, as the learning process must be unrolled to calculate gradients with respect to the initial parameters \citep{antoniou2018train}. 
In practice, the learning process is often truncated to a small number of gradient steps, but 
has been shown to have a short-horizon bias \citep{wu2018understanding}, causing it to underperform. 

In this work we explore OOV word embedding learning using \textit{Leap} \citep{flennerhag2018transferring}, a meta-learning framework which takes into consideration the entire learning trajectory, not only the beginning and end points.
Each task is associated with a loss surface over the model's parameters on which the learning process travels, and the aim is to minimize the expected length of this process across tasks. 
Leap also does not require backpropagation through the learning process, allowing it to adapt over a larger number of gradient steps and thus not suffering from the short-horizon bias that MAML is prone to.

We conduct an intrinsic evaluation of MAML and Leap on the
dataset of \citet{chimeras} which simulates OOV words by combining the contexts of two semantically similar words to form a 'chimera'. We find that Leap performs better than MAML at adapting model parameters to a new corpus. We also conduct an extrinsic evaluation on NER in the biomedical domain where the results are comparable to MAML, without improving in most cases on a random embedding baseline. Finally, we examine which contexts are more beneficial to learn an embedding from, and note that the contexts from which an embedding is learned  matters more than the meta-learning method employed.

\section{Meta-Learning}

Meta-learning algorithms aim to capture knowledge across a variety of learning tasks such that fine-tuning a model on a specific task both avoids overfitting and leads to good performance.
Approaches include learning a similarity function by which to cluster and classify data points \citep{vinyals2016matching, snell2017prototypical}; learning an update rule for neural network optimization \citep{Ravi2017OptimizationAA}; and learning an initialization from which to fine-tune a model. We consider algorithms for the latter.

Formally, we consider task $\mathcal{T}$ to consist of a dataset $D_\mathcal{T}$ of labelled examples $(x,y)$ and loss function $\mathcal{L}_\mathcal{T}(\theta) \rightarrow \mathbb{R}$ which maps a model's parameters $\theta$ to a real-valued loss. For batch gradient descent this loss is constant for a given set of parameters, however for stochastic gradient descent it depends on the sampled examples. During training, tasks $\mathcal{T} \sim p(\mathcal{T})$ are sampled from a distribution $p(\mathcal{T})$ over the tasks the model should be able to adapt to. The aim is then to train the model's parameters such that they capture features common to all tasks in $p(\mathcal{T})$, and thus are a promising initialization for any of these tasks. Below we describe the approaches taken by MAML and Leap.

\subsection{MAML}

During training with MAML, for each task $\mathcal{T}$ the model's parameters are updated from an initialization $\theta$ to $\theta_\mathcal{T}^{K}$ through $K$ gradient steps according to $\mathcal{L}_\mathcal{T}$. The final parameters $\theta_\mathcal{T}^{K}$ are then used to calculate the final task losses, and the original parameters $\theta$ are updated to minimize their sum. 
This meta-objective is given below in equation \ref{maml-obj}, where $u_\mathcal{T}(\theta) = \theta - \alpha \nabla_{\theta} \mathcal{L}_\mathcal{T} (\theta)$ is a single gradient step:
\begin{equation}\label{maml-obj}
  \min_\theta \sum_{\mathcal{T} \sim p(\mathcal{T})} \mathcal{L}_{\mathcal{T}}(\theta_\mathcal{T}^{K}) = 
  \sum_{\mathcal{T} \sim p(\mathcal{T})} \mathcal{L}_{\mathcal{T}}( u^K_\mathcal{T}(\theta) )
\end{equation}
The final task losses $\mathcal{L}_{\mathcal{T}}(\theta_\mathcal{T}^{K})$ are usually computed using examples held out from training $\theta_\mathcal{T}^{K}$ to simulate a testing loss. The meta-optimization is then performed by backpropagating with respect to the original parameters, $\theta$, rather than the trained parameters $\theta_\mathcal{T}^{K}$. This aims to optimize $\theta$ such that a small number of gradient steps, $K$, on a particular task produces a low testing loss. Algorithm~\ref{maml-task} below gives the overview of the training.

\begin{algorithm}[]
\SetAlgoLined
\SetKwFor{RepTimes}{repeat}{times}{end}
\KwIn{$p(\mathcal{T})$: a distribution over tasks}
\KwIn{$\alpha, \beta$: step size parameters}
define $u_\mathcal{T}(\theta) = \theta - \alpha \nabla_{\theta} \mathcal{L}_\mathcal{T} (\theta)$\;
initialize model parameters $\theta$\;
\While{not done}{
sample batch $\mathcal{B}$ of tasks $\mathcal{T} \sim p(\mathcal{T})$\;
\ForAll{$\mathcal{T} \in \mathcal{B}$}{
 $\theta_\mathcal{T}^{K} \leftarrow u^K_\mathcal{T}(\theta)$\;
}
$\theta \leftarrow \theta - \beta \nabla_{\theta} \sum_{\mathcal{T} \in \mathcal{B}} \mathcal{L}_{\mathcal{T}}(\theta_\mathcal{T}^{K})$\;
}
\caption{MAML}
\label{maml-task}
\end{algorithm}

\begin{algorithm*}[]
\SetAlgoLined
\SetKwFor{RepTimes}{repeat}{times}{end}
\KwIn{$p(\mathcal{T})$: a distribution over tasks}
\KwIn{$\alpha, \beta$: step size parameters}
initialize model parameters $\theta$\;
\While{not done}{
$\nabla \bar{F} \leftarrow 0$\; 
sample batch $\mathcal{B}$ of tasks $\mathcal{T} \sim p(\mathcal{T})$\;
\ForAll{$\mathcal{T}$}{
 $\theta_\mathcal{T}^0 \leftarrow \theta$\;
 $\psi_\mathcal{T}^0 \leftarrow \theta$\;
 \ForAll{$i \in \{ 0, ..., K-1 \}$}{
  $\theta_\mathcal{T}^{i+1} \leftarrow \theta_\mathcal{T}^i - \alpha \nabla_{\theta_\mathcal{T}^i} \mathcal{L}_{\mathcal{T}} (\theta_\mathcal{T}^i)$\;
  
  $\psi_\mathcal{T}^{i+1} \leftarrow \psi_\mathcal{T}^i - \alpha \nabla_{\psi_\mathcal{T}^i} \mathcal{L}_{\mathcal{T}} (\psi_\mathcal{T}^i)$\;
  
  $\nabla \bar{F} \leftarrow \nabla \bar{F} + \frac{(\mathcal{L}_\mathcal{T}(\theta_\mathcal{T}^{i}) - \mathcal{L}_\mathcal{T}(\psi_\mathcal{T}^{i+1})) \nabla_{\theta_\mathcal{T}^i} \mathcal{L}_{\mathcal{T}} (\theta_{\mathcal{T}}^i) + (\theta_\mathcal{T}^{i} - \psi_\mathcal{T}^{i+1})}{\Vert \bar{\gamma}_{\mathcal{T}}^{i+1} - \gamma_{\mathcal{T}}^{i} \Vert_2}$ \;
 }
}
$\theta \leftarrow \theta - \frac{\beta}{|\mathcal{B}|} \nabla \bar{F}$\;
}
\caption{Leap}
\label{leap-task}
\end{algorithm*}

Computing $\nabla_{\theta} \sum_{\mathcal{T} \in \mathcal{B}} \mathcal{L}_{\mathcal{T}}(\theta_\mathcal{T}^{K})$ requires backpropagation through the learning process for each task $\mathcal{T} \in \mathcal{B}$. Gradients are computed by backpropagation through $K+1$ different instances of the model's parameters, which can cause both exploding and diminishing gradient problems, and becomes
unstable for large $K$ \citep{antoniou2018train}. However, truncating the learning process with a small $K$ results in a short-horizon bias \citep{wu2018understanding}, where the learned parameters adapt poorly to tasks over a number of gradient steps larger than $K$.
To extend to more steps, a first order approximation of MAML 
has been shown to achieve similar performance \citep{nichol2018firstorder, finn2017modelagnostic}. 
However, this still considers only the initial and the final parameters and loss, and for larger $K$ the intermediate steps become more significant.

\subsection{Leap}

In Leap the learning process is viewed as a path along a loss surface $\mathcal{L}$, traversed by $K$ gradient steps from initial to final parameters. The intuition behind Leap is that geometrical similarities between learning processes associated with different tasks can be exploited for transfer learning. In particular, Leap seeks to find an initialization that reduces the expected length of learning processes. 

Following \citet{flennerhag2018transferring}, we consider the learning process to be
a sequence of discrete points $\{\gamma^i\}_{i=0}^{K}$ with $\gamma^i = (\theta^i, \mathcal{L}(\theta^i))$ corresponding to $K$ gradient updates, and as such we consider the learning process to be the shortest path passing through these points. The length of a learning process is then approximated as the cumulative chordal distance of the path from initial parameters $\theta = \theta^0$ to final parameters $\theta^K$.
The cumulative chordal distance approximates the length of the arc passing through the points $\{\gamma^i\}_{i=0}^{K}$, and is key to minimizing the length of the learning process rather than simply moving the initial parameters towards the final parameters:
\begin{equation}\label{cum-chord}
  d(\theta; \mathcal{L}) = \sum_{i=0}^{K-1} \Vert \gamma^{i+1} - \gamma^{i} \Vert_2
\end{equation}
Considering again a distribution of tasks $p(\mathcal{T})$, an initialization $\theta$ is associated with an expected learning process length $\mathbb{E}_{\mathcal{T} \sim p(\mathcal{T})}[d(\theta; \mathcal{L}_\mathcal{T})]$. When dealing with complex non-convex loss surfaces, minimizing only the expected learning process length makes no guarantees about the final loss $\mathcal{L}(\theta^K)$ and thus may inadvertently find final parameters $\theta^K$ with lower performance. Note that MAML takes this into account directly in its objective by optimizing for the final loss on a separate held-out set. Leap instead enforces this as part of its meta-objective by requiring that the optimized initialization $\theta$ must not converge to a higher loss than a baseline initialization $\psi = \psi^0$ for all tasks $\mathcal{T}$. That is, $\mathcal{L}_\mathcal{T}(\theta^K) \leq \mathcal{L}_\mathcal{T}(\psi^K)$ assuming convergence after $K$ gradient steps. For this purpose, Leap defines an
objective which optimizes the initialization for expected learning process length only along the task-specific learning processes originating from a baseline initialization. This ensures that the learning processes originating from the optimized initialization will have no greater final loss than their counterpart originating from the baseline initialization, given that both consist of $K$ gradient steps. The objective is given in equation \ref{pull-forward}, where points $\{\gamma_{\mathcal{T}}^i\}_{i=0}^{K}$ lie along the learning process for task $\mathcal{T}$ originating from the optimized initialization $\theta$, while points $\{\bar{\gamma}_{\mathcal{T}}^i\}_{i=0}^{K}$ lie along the respective learning process originating from the fixed baseline initialization $\psi$.
\begin{equation}\label{pull-forward}
\begin{aligned}
  \bar{d}(\theta; \mathcal{L}_\mathcal{T}, \psi) &= \sum_{i=0}^{K-1} \Vert \bar{\gamma}_{\mathcal{T}}^{i+1} - \gamma_{\mathcal{T}}^{i} \Vert_2 \\
  \min_{\theta} \bar{F}(\theta; \psi) &= \mathbb{E}_{\mathcal{T} \sim p(\mathcal{T})}[\bar{d}(\theta; \mathcal{L}_\mathcal{T}, \psi)]
\end{aligned}
\end{equation}
Gradient descent on the 
objective $\bar{F}$ with respect to $\theta$ pulls the parameters $\theta^i$ towards $\psi^{i+1}$. Parameters $\theta$ are initialized to be equal to $\psi$ such that each gradient descent update pulls $\theta$ forward along the learning processes originating from it.

Algorithm~\ref{leap-task} describes the training with Leap. Tasks $\mathcal{T} \sim p(\mathcal{T})$ are sampled from the distribution $p(\mathcal{T})$ and the model's parameters are updated to $\theta_\mathcal{T}^{K}$ through $K$ gradient steps for each task $\mathcal{T}$ as in MAML (lines 1-10; the learning trajectory is expanded in lines 8, 9 and 10, as it is needed in Leap). The 
gradient $\nabla \bar{F}$ is incrementally computed at each point $(\theta_\mathcal{T}^i, \mathcal{L}_\mathcal{T}(\theta_\mathcal{T}^i))$ during the task training (line 11). $\nabla \bar{F}$ is always evaluated at $\theta = \psi$, with the update term on line 11 pulling $\theta^i$ towards $\psi^{i+1}=\theta^{i+1}$. This is performed in order to take into account each point in the learning trajectory, as opposed to the start and end points in MAML.
The initialization $\theta$ is then updated according to the accumulated gradient $\nabla \bar{F}$ (line 14), and the algorithm continues with $\psi$ set to the updated $\theta$. This is done implicitly when $\theta$ is updated, which ensures that any future $\theta$ improves the task loss over the one already obtained, instead of just over the initial random initialization.

\section{Leap for OOV Embedding Learning}

In this section we describe our application of Leap to learning embeddings for OOV words. This technique is generally applicable to any word embedding regression function $H_\theta$ trainable by gradient descent, though in our experiments we use the HiCE architecture \citep{hu2019fewshot}.

The regression function $H_\theta$ is trained to predict a word's embedding given K contexts, and possibly morphological information, with words and their contexts sampled from some large training corpus $\mathcal{D}_T$. Each word (either a training target or in a context) is represented as a pre-trained embedding in the input to $H_\theta$. 
We note that $H_\theta$ can be trained only on words with sufficient occurrences in $\mathcal{D}_T$ such that their pre-trained embeddings can be accurately learned. The trained $H_\theta$ can then be used to infer the embeddings for OOV words, for which we do not have a pre-trained embedding. However, OOV words often form part of domain-specific vocabularies, the semantics of which are not captured by word embeddings trained on generic large corpora \citep{kameswara-sarma-etal-2018-domain}. To counteract this, we adapt the trained parameters $\theta_T$ of the word embedding regression function to the domain in which we infer the OOV embeddings.

When adapting $\theta_T$ we consider both $\mathcal{D}_T$, and the corpus on which we wish to infer OOV embeddings, $\mathcal{D}_N$. We wish to transfer the knowledge encoded in $\theta_T$ to the task of predicting OOV embeddings for words in $\mathcal{D}_N$. One approach would be to simply fine-tune $\theta_T$ on $\mathcal{D}_N$; that is, sample words and their contexts from $\mathcal{D}_N$ which have their pre-trained embeddings from $\mathcal{D}_T$ known and train $H_\theta$ as before on these words. This ignores that $\mathcal{D}_N$ is much smaller than a corpus usually used for word embedding training, and direct training on it is likely to lead to overfitting to the corpus rather than adapting to its domain, which in turn hurts the quality of inferred OOV embeddings, an instance of catastrophic forgetting \citep{FRENCH1999128}. Instead of fine-tuning, meta-learning algorithms can be applied. \citet{hu2019fewshot} use MAML, and we extend this work with the use of Leap.

We adapt $\theta_T$ by applying Leap across two tasks; inferring OOV words in $\mathcal{D}_T$ and $\mathcal{D}_N$. Optimizing the 
objective in equation \ref{oov-pull-forward} moves the adapted parameters $\theta$ to minimize the lengths of the two corresponding learning processes. The loss function $\mathcal{L}(\cdot | \mathcal{D})$ used throughout is the cosine distance between predicted and pre-trained embeddings.
\begin{equation}\label{oov-pull-forward}
\begin{aligned}
  \min_{\theta}
  \sum_{C \in \{ T, N \}} \bar{d}(\theta; \mathcal{L}(\cdot | \mathcal{D}_C), \theta_T)
\end{aligned}
\end{equation}
This pulls the model parameters $\theta$ along the learning processes for $\mathcal{D}_T$ and $\mathcal{D}_N$ originating from the pre-trained parameters $\theta_T$. The learning process for $\mathcal{D}_T$ should already be short, as $\theta_T$ is trained to convergence on $\mathcal{D}_T$. Optimization of the pull-forward objective must naturally pull the parameters away from this point of convergence to minimize the length of the learning process for $\mathcal{D}_N$. However, as each task is weighted equally, the model parameters cannot move towards the convergence point for $\mathcal{D}_N$ if this results in a 
larger divergence from the convergence point for $\mathcal{D}_T$. It is thus necessary for the parameters to move into areas which encode knowledge sharing between $\mathcal{D}_T$ and $\mathcal{D}_N$. This reduces overfitting on $\mathcal{D}_N$ by ensuring $\theta$ does not move too far from the convergence point for $\mathcal{D}_T$.

While each word's embedding is inferred from only a few contexts, the total number of words available for training in $\mathcal{D}_N$ is large enough so $\theta$ can be adapted over a larger number of gradient steps. We posit that this, in conjunction with knowledge sharing that considers the entire learning trajectory rather than just the beginning and end points, results in higher quality OOV embeddings than those that can be obtained with MAML.

\section{Experiments}

\subsection{Intrinsic Evaluation}
\label{intrinsic-eval}

%
%
%
    
    
    
    
      
    
    

\begin{table*}[t]
  \centering
\begin{tabular}{|l|l|l|l|}
  \cline{2-4}
 \multicolumn{1}{c|}{} & 2-shot & 4-shot & 6-shot  \\
 \cline{2-4} 
 \noalign{\vskip\doublerulesep
    \vskip-\arrayrulewidth}
\hline
fasttext & 0.1775 & 0.1738 & 0.1294  \\
\hline
additive & 0.3376 & 0.3624 & 0.4080  \\
\hline
a la carte &  0.3634 & 0.3844 & 0.3941 \\
\hline
nonce2vec & 0.3320 & 0.3668 & 0.3890 \\
\hline
\hline
HiCE & 0.3611 $\pm$ 0.0054 & 0.3882 $\pm$ 0.0049 & 0.4134 $\pm$ 0.0043 \\
\hline
+MAML & 0.3574 $\pm$ 0.0058 & 0.3901 $\pm$ 0.0072 & 0.4169 $\pm$ 0.0057 \\
\hline
+Leap & \textbf{0.3695 $\pm$ 0.0022} & \textbf{0.4022 $\pm$ 0.0043} & \textbf{0.4262 $\pm$ 0.0051} \\
\hline \hline
pre-trained & 0.4173 & 0.4367 & 0.4410 \\
\hline
\end{tabular}
\caption{Average Spearman's correlations for each k-shot case and method. Resulting of HiCE are given with 95\% confidence intervals. Bold indicates the best results.}
\label{chimera-eval}
\end{table*}

To obtain an intrinsic evaluation of the methods proposed, we require a dataset that simulates the natural occurrences of OOV words in a real-world setting, and defines a notion for evaluating how close an embedding is to representing an OOV word's meaning. Following \citet{hu2019fewshot}, we use the 'Chimera' dataset for evaluation, a popular benchmark dataset for OOV words. 

The Chimera dataset \citep{chimeras} is constructed specifically to simulate unseen words occurring naturally in text. Each unseen word is a \textit{chimera}, which is a novel concept created by combining two related but distinct concepts; for example a gorilla and a bear. In total there are 33 chimeras, generated by first taking a base concept, called a \textit{pivot}, and matching this with a compatible concept by traversing a list of terms ranked by similarity to the pivot. In the case of the 'gorilla/bear' chimera, the pivot is the gorilla and the compatible term is the bear. Each chimera is then associated with passages of 2, 4 and 6 sentences, 
with half containing the pivot and half containing the compatible concept. The occurrences of the pivot and the compatible term are replaced with a nonce word that represents the chimera; for example 'mohalk'. 

Each passage is then annotated by human subjects with similarity scores between the nonce word and six different words, called \textit{probes}, specific to the chimera. These similarity scores 
are then averaged across human subjects, resulting in six scores for each passage indicating the similarity between the chimera and each probe word. 

For each of the 2-shot, 4-shot and 6-shot cases the chimera's embedding is inferred given each sentence (shot) as a context and the pivot's character sequence. Following \citet{chimeras} we measure the performance of the embeddings inferred by looking at their cosine similarity to the probe embeddings and calculating the Spearman correlation to the similarity judgements by the human subjects. 

Throughout all experiments HiCE is trained on WikiText-103 \citep{merity2016pointer} with pre-trained embeddings provided by SkipGram \citep{mikolov2013distributed} on the same corpus. Where MAML and Leap are used, word embeddings are adapted to the Chimera dataset. Both approaches were implemented using PyTorch \citep{paszke2017automatic} and the code will become publicly available. To implement MAML we use \texttt{higher} \citep{grefenstette2020generalized}, a PyTorch library for higher order optimization such as backpropagation through gradient descent updates, as is required for MAML. This allows us to compute the second order gradients that are required for MAML, rather than using a first order approximation. The learning rates $\alpha$ and $\beta$ for each of MAML and Leap were chosen based on each algorithm's stability during training; for MAML we used $\alpha=5 \times 10^{-4}, \beta=1 \times 10^{-5}$ and for Leap we used $\alpha=5 \times 10^{-4}, \beta=1 \times 10^{-4}$. These learning rates are in line with the publicly available code of \citet{hu2019fewshot}. The number of gradient steps used for adaptation with MAML was 4, while with Leap we increased this to 64, taking advantage of its ability to train tasks over longer horizons.\footnote{The code used in our experiments is available here: \url{http://github.com/Gordonbuck/ml-oov-we}}


Table \ref{chimera-eval} gives the average Spearman correlations for HiCE, its combination with MAML as proposed by \citet{hu2019fewshot}, and its combination with Leap as proposed in this work, for the 2-shot, 4-shot and 6-shot cases. We also include the results for related works taken from their corresponding papers. These include fasttext \citep{bojanowski2016enriching}; the additive method (a simple averaging of context word embeddings) \citep{chimeras}; a la carte (a linear transformation-based modification of the averaging method) \citep{khodak2018la}; and nonce2vec (a modification of the Word2Vec algorithm for few-shot learning) \citep{herbelot2017highrisk}. We also give the scores obtained when using the pivot's pre-trained embedding as the chimera's embedding to indicate a ceiling, following \citet{hu2019fewshot}. However, we would expect the OOV embedding to differ from the pivot's pre-trained embedding, since the semantics of a chimera are a combination of the pivot and compatible concept.

HiCE+Leap achieves the best results across all k-shot settings in our experiments. The results for HiCE, HiCE+MAML and HiCE+Leap are all obtained by averaging the results over 10 different random seeds, and we give a 95\% confidence interval for each. We take this approach to highlight the known instability of training MAML across random seeds \citep{antoniou2018train}, even with no hyperparameter changes. Leap consistently performs better than MAML and with a lower variance; in all cases, the average Spearman's correlation for MAML lies outside of the confidence interval range given for Leap. We also see that due to MAML's instability it can actually lower the performance of pre-trained HiCE in the 2-shot case. Outside of HiCE-based methods, a la carte performs best in the 2- and 4-shot cases, and additive performs best in the 6-shot case. These scores similarly lie outside of confidence interval ranges for HiCE+Leap, which overall performs best in each case.

\citet{hu2019fewshot} reported 0.3781,  0.4053 and 0.4307 with HiCE+MAML in the in the 2-, 4- and 6-shot cases respectively, but did not report experiments with multiple random seeds. While were able to obtain similar results for HiCE+MAML in some of our experiments, they were outside the confidence intervals we obtained, illustrating the relative instability in training with MAML \cite{antoniou2018train}. The results of \citet{hu2019fewshot} are also lower than the highest results we obtained with HiCE+Leap
(0.3896, 0.4116 and 0.4395 for 2-, 4- and 6-shot).



\subsection{Extrinsic Evaluation}

To gauge the quality of the OOV embeddings for downstream tasks we evaluate their performance when applied to NER. For this purpose we use the JNLPBA 2004 Bio-Entity Recognition Task dataset \citep{collier-kim-2004-introduction}. We choose this dataset as the biomedical domain differs significantly from the domain of Wikipedia that HiCE is pre-trained on, and contains many OOV technical terms. \citet{hu2019fewshot} use this dataset also but did not provide their datasplits; thus. while we were able to confirm their results, we cannot compare against them directly.

The JNLPBA dataset is constructed from 2000 abstracts for training and 404 abstracts for testing, each extracted from a bibliographic database of biomedical information and hand annotated with 36 classes corresponding to chemical classifications. These classifications are simplified into 5 classes for the purpose of the bio-entity recognition task; \textit{protein}, \textit{DNA}, \textit{RNA}, \textit{cell-line} and \textit{cell-type}. In total there are 18546 training and 3856 test sentences.

Following \citet{hu2019fewshot}, HiCE is trained on the WikiText-103 corpus, and we adapt word embeddings to the biomedical domain by using the JNLPBA dataset as a corpus. Contrary to the Chimera dataset, we consider contexts at the abstract level rather than only the sentence level. We train different embeddings for OOV words for each of the 2-shot, 6-shot and 10-shot cases, considering only those OOV words with 2, 6 or 10 occurrences or more respectively. In total we infer embeddings for 4643 OOV distinct words (types) in the 2-shot case, 1310 in the 6-shot case and 702 in the 10-shot case.\footnote{\citet{hu2019fewshot} do not distinguish between different number of shots/contexts per word in their results.} 
The contexts used to infer the embedding of a word are chosen at random from the contexts that word appears. The inferred embeddings, alongside the embeddings for in-vocabulary words, are used as input to train the LSTM-CRF architecture of \citet{lample2016neural}. 

For each of the k-shot cases a separate test set is created by subsampling, such that the respective test set contains only those sentences with an OOV word whose embeddings has been inferred. This ensures that the test sets focus on the quality of the inferred OOV embeddings. For the 2-shot case there are 2876 test sentences; 2451 for 6-shot; and 2134 for 10-shot. The results for each k-shot setting are given in Table \ref{jnlpba-eval}, reported in micro-averaged F1 score. The results obtained using random OOV embeddings as input are given as a baseline. 

\begin{table}[t]
  \centering
\begin{tabular}{|l|l|l|l|}
  \cline{2-4}
 \multicolumn{1}{c|}{} & 2-shot & 6-shot & 10-shot  \\
 \cline{2-4} 
 \noalign{\vskip\doublerulesep
    \vskip-\arrayrulewidth}
 \hline
random & \textbf{0.7226} & 0.7206 & 0.7209 \\
\hline \hline
HiCE & 0.7116 & 0.7213 & 0.7232  \\
\hline
+MAML & 0.7135 & 0.7232 & 0.7269 \\
\hline
+Leap & 0.7141 & \textbf{0.7256} & \textbf{0.7282 $\dagger$} \\
\hline
\end{tabular}
\caption{Micro-averaged F1 score for each of the 2-shot, 6-shot and 10-shot settings. Bold indicates the best results and ($\dagger$) indicates the result is better than random embeddings at a 0.05 significance level.}
\label{jnlpba-eval}
\end{table}

Results are marginally improved with any of the proposed methods for the 6-shot and 10-shot cases, with HiCE+Leap producing the best results. 
We perform a paired t-test on each pair of results within a k-shot case. However, we do not find the differences between methods to be significant, with only HiCE+Leap in the 10-shot case performing significantly better than random embeddings, at a 0.05 significance level.

We find that performance
increases 
across all methods as we use more contexts. However, for the 2-shot case we find that results are lower than if random embeddings are used. With fewer contexts the quality of the learned OOV embeddings is naturally lower, and inaccuracies in the embeddings add noise which hurts the performance of the downstream NER tagger. This highlights the need for a sufficient number of occurrences to effectively learn word embeddings, even with models specifically designed to handle the lack of data, and that using inaccurate embeddings can lower the performance of the entire downstream system. 
Our findings corroborate contemporary research which suggests that random embeddings can perform comparably to pre-trained and contextual embeddings on benchmark tasks \citep{arora2020contextual}. 

\subsection{Informative Contexts}

\begin{table*}[ht]
\centering
 \begin{tabular}{|| p{2cm} | p{3cm} | p{4cm} | p{4cm} | p{1cm} ||} 
 \hline
 chimera & probes & shot 1 & shot 2 & score \\
 \hline\hline
 refrigerator / closet & cupboard, basement, mixer, dishwasher, ladle, boat & 23144 security materials that may adversely affect hpc components shall not be stored in the \_\_\_ or freezers & if there is any difference from the first there are fewer of those \_\_\_ & -0.68 \\ 
 \hline
 broccoli / spinach & celery, radish, grape, salamander, budgie, pot & all manner of produce fill the fields including exotic vegetables like pumpkins spring onions \_\_\_ and asparagus & discard any bruised or yellow leaves and soak remaining leaves in a basin of cold water with \_\_\_ & 0.98 \\
 \hline
 \hline
 drum / tuba & bagpipe, harmonica, whistle, shotgun, bear, bouquet & i could nt yet hear the \_\_\_ that told you where to go on any particular day & is probably at mirage poker room income from week he borrowed was swimming \_\_\_ & -0.63 \\
 \hline
 drum / tuba & bagpipe, harmonica, whistle, shotgun, bear, bouquet & is there a better way to start a record or a song than with a thumpin \_\_\_ intro & they add considerably to the tone of the \_\_\_ however when used on their low notes & 0.91 \\
 \hline
\end{tabular}
\caption{Each section gives a higher and lower performing 2-shot passage with its average Spearman correlation across all methods and random seeds. 
The nonce word is replaced by '\_\_\_'.}
\label{chimera-rank}
\end{table*}

Apart from comparing different meta-learning approaches for word embedding learning,
we seek to find which contexts are invariably most informative to a word's meaning. For this purpose, we return to the Chimera dataset. Considering the 2-shot case, we rank passages by the performance of the chimera embeddings inferred from them. That is, we infer the chimera embeddings, obtain the cosine similarities against the probe embeddings, and calculate the Spearman correlation against the human scores.
We then calculate the pairwise Spearman correlation between these rankings across random seeds and methods experimented with in this paper, i.e.\ HiCE, HiCE with MAML, and HiCE with Leap. The average Spearman correlation is $0.89 \pm 0.0047$ for a 95\% confidence interval, indicating that the rankings of passages are largely similar across methods. 
Thus we conclude that the contexts which each method finds most useful to infer a chimera's meaning are largely invariable.

Ordering passages by their cumulative rank across 
methods, we observe that passages which consistently perform poorly are composed of sentences with more ambiguity and fewer content words. The top section in table \ref{chimera-rank} gives the lowest and highest performing passages as examples. The lowest performing passage contains little in the way of content words indicating the meaning of the chimera 'refrigerator/closet' besides the presence of 'freezers', and the second sentence is highly ambiguous. Naturally, human annotators would also struggle to pinpoint the semantic meaning of this chimera based on the two sentences given. In contrast the highest performing passage very clearly relates to organic produce and cooking, providing far more hints as to the semantic meaning of the chimera 'broccoli/spinach'.

We also look at which passages perform best for a given chimera. The bottom section in table \ref{chimera-rank} gives the lowest and highest performing passages for the 'drum/tuba' chimera. We observe the same trend within the chimera, with the lowest performing passage consisting of more ambiguous sentences, while the highest performing passage contains many content words which hint at the semantics of the chimera, such as 'song'. These two passages differ greatly in their performance, with the lowest passage averaging a Spearman correlation of $-0.63$ while the highest passage achieves an average of $0.91$. However, if we look across methods the difference in performance for each of these passages is no higher than $0.2$. This suggests that one of the most important factors to inferring an OOV word's embedding is the choice of contexts, perhaps more so than the meta-learning employed. To quantify this further, we calculate the Spearman correlation between the proportion of content words in a passage and its score, which we find to be $0.12 \pm 0.0083$ for a 95\% confidence interval. We do this by using a standard part-of-speech tagger \citep{spacy2} on each passage; taking all nouns, adjectives, verbs and adverbs to be content words. While the correlation is weak, it is significant at a 95\% confidence interval and further suggests that context informativeness is a suitable future area of work.

\section{Conclusion}

We investigated the use of meta-learning for few-shot learning of OOV word embeddings. We built on the work of \citet{hu2019fewshot}, formulating OOV word embedding learning as a few-shot regression problem and training their proposed architecture HiCE to predict OOV embeddings given $K$ contexts and morphological features. We proposed the use of Leap as a meta-learning algorithm to adapt HiCE to a new semantic domain and compared it to the popular MAML as used by \citet{hu2019fewshot}. Experiments on a benchmark dataset show that Leap is more stable and achieves comparably higher performance than MAML in the context of OOV embedding learning. Further experimentation shows that there is little variation in which contexts perform well across both random seeds and meta-learning approaches, and a qualitative analysis indicates that performance is lower on ambiguous sentences with fewer content words. 
Our findings suggest a future avenue of work which focuses on the selection of contexts from which to learn an OOV embedding, such as prioritising contexts based on a notion of informativeness.

\bibliography{anthology,emnlp2020}
\bibliographystyle{acl_natbib}

\end{document}